# Research on facial expression recognition based on Multimodal data fusion and neural network

Han Yi[1,2]  Wang Xubin[2]  Lu Zhengyu[2*]

1. School of Mechanical Science and Engineering, Huazhong University of Science and Technology, Wuhan, China

2. Anyang Institute of Technology Department of Computer Science and Information Engineering, Anyang China

**Corresponding author:** Lu Zhengyu ,Anyang Institute of Technology Department of Computer Science and Information Engineering, Anyang China

**Abstract**: Facial expression recognition is a challenging task when neural network is applied to pattern recognition. Most of the current recognition research is based on single source facial data, which generally has the disadvantages of low accuracy and low robustness. In this paper, a neural network algorithm of facial expression recognition based on multimodal data fusion is proposed. The algorithm is based on the multimodal data, and it takes the facial image, the histogram of oriented gradient of the image and the facial landmarks as the input, and establishes CNN, LNN and HNN three sub neural networks to extract data features, using multimodal data feature fusion mechanism to improve the accuracy of facial expression recognition. Experimental results show that, benefiting by the complementarity of multimodal data, the algorithm has a great improvement in accuracy, robustness and detection speed compared with the traditional facial expression recognition algorithm. Especially in the case of partial occlusion, illumination and head posture transformation, the algorithm also shows a high confidence.

## 0.Introduction

Facial expression recognition model is an important method of human-computer interaction and pattern recognition, which has been widely studied in many fields [1]. Automatic facial expression recognition by computer makes many new applications come true. For the task of facial expression recognition, many traditional recognition methods have been used. For example, Sariyanidi et al. [2] use Gabor wavelet and SVM to recognize facial expression, extract features through filter and send them to SVM for classification; Lanitis et al. [3] use geometric feature method to recognize facial expression, and calculate the relative position by marking five official landmarks, which are highly dependent on the previous manual feature extraction. The result is not good enough because of human interference. At present, the use of deep learning and convolutional neural network is better than other methods, and has achieved good results. For example, Tang [4] uses convolutional neural network and SVM for expression recognition, and the results are better than traditional methods; Agrawal et al. [5] propose two novel convolutional neural network architectures through the evaluation of different convolutional neural network parameters, and achieves an accuracy of 65% on the data set; Li et al Human [6] uses the classic lenet-5 convolutional neural network to recognize facial expression, combines the low-level features extracted from the network structure with the high-level features to construct a classifier, and achieves good results.

The human visual system has a special mechanism to recognize facial expressions and perceive them mainly through the whole, part and structure [7]. When perceiving expression, it is often recognized by multi-dimensional features rather than single source information. At present, most facial expression recognition models are only based on single source data, and do not make full use of the implicit information between the data. The facial image contains a variety of multimodal data information, such as head posture, facial expression and so on; or implicitly, it needs to process the histogram of oriented gradient of the image, the expression of abstract semantic features and so on. The information that can be expressed by using single source data is limited, but the Multimodal data fusion of the face and the cooperation between the data can maximize the information required by the task and improve the accuracy of the algorithm.

The strong nonlinear expression ability of neural network makes it a good way to extract features, and the excellent feature expression ability makes it have a broad application in the field of data fusion. For example, Chen et al. [8] propose a data fusion model based on deep convolution neural network for fault diagnosis. By installing sensors in different positions to collect horizontal and vertical vibration signal data, the sensor data can describe the fault characteristics more comprehensively and accurately. The results show that the detection model has better detection performance and scalability by introducing the data fusion method of deep convolution neural network; Du et al. [9] propose a deep learning framework based on CNN, GRU (Gated Recurrent Unit) and adaptive multimodal joint model for short-term traffic flow prediction, which uses deep learning model in the process of feature extraction and fusion of multimodal data.

In this paper, a neural network driven by multimodal data is designed for facial expression recognition based on a number of multimodal data, namely facial image, facial landmarks and histogram of oriented gradient of facial image. The deep learning model is used in feature extraction and feature fusion of multimodal data, and the nonlinear mapping ability of neural network is exploited for feature extraction. At the same time, attention mechanism is introduced to improve the performance of the model.

# 1. Related Work

## 1.1 Multimodal Data

In the traditional research of facial expression recognition, convolution neural network and other methods are often used to construct a learner to extract features from the image information to recognize facial expression. This kind of model can generally learn a large number of single expression features in the input image, but in special cases, such as facial occlusion, illumination changes, the accuracy of the model will decline or even can not be predicted.

The model based on multimodal data fusion can improve this situation. When the model can not get complete facial features, it can still obtain them from other sources for prediction. The neural network designed in this paper is driven by three multimodal data: face image, facial landmarks and histogram of oriented gradient (HOG). As shown in Fig. 1, the neural network can output the prediction results through the information of image itself and that hidden in the image.

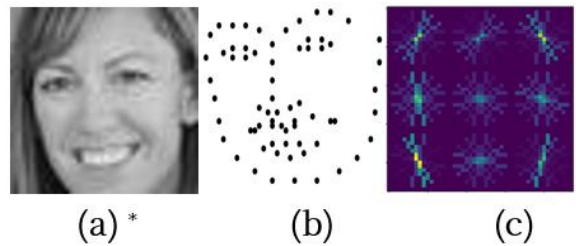

Fig. 1. Three multimodal  facial data，（a）Facial Image；（b）Facial landmarks；（c）HOG Feature.

## 1.2 Design of Neural Network

A variety of ways can be used in the neural network to process multimodal  data, such as fusion based on feature layer, which means the feature map extracted from each sub network is fused, and then the full connection layer or classifier outputs the final result; or fusion based on classification layer, in other words, the feature is extracted and classified from the sub network, and then the classification results are fused to output the final result.

Compared with the fusion method based on classification layer, the fusion method based on feature layer is more in line with the requirements of the neural network. The former fuses the classification results of the sub network and then classify the fusion data, which will lose part of the information when classifying in the fully connected layer of the sub network; and the latter classifies by fusing the features extracted by each sub network, which can retain the information of each input feature to the greatest extent, making the final classifier work more perfect.

Therefore, for the input of the three multimodal data, the neural network will first establish a sub network to perform their own extraction work, then fuse the feature map, and ultimately output the final result by the classifier.

## 1.3 Related Data Sets

This work will be mainly based on Fer2013 data set and CK + data set, as shown in Fig. 2.

Fer2013 was created by Pierre Luc carrier et al. [10] and was introduced into the expression recognition challenge of ICML2013. The dataset contains 35887 images, including 28709 images for training (OTD), 3589 images for public test (PTD) and 3589 images for final test (FTD). Each category accounts are shown in Fig. 2 (b). There are great differences in the age and posture of faces, including not only real face images, but also cartoon face images, which are very challenging.

CK + is a general facial expression data set [11], which is suitable for facial expression recognition. It is based on the Cohn Kanda data set and contains 593 image sequences of 123 participants. In the acquisition process, the number of image sequences for each tester is not the same, at least 10 frames and up to 60 frames. The percentage of each category after processing is shown in Fig. 2 (d).

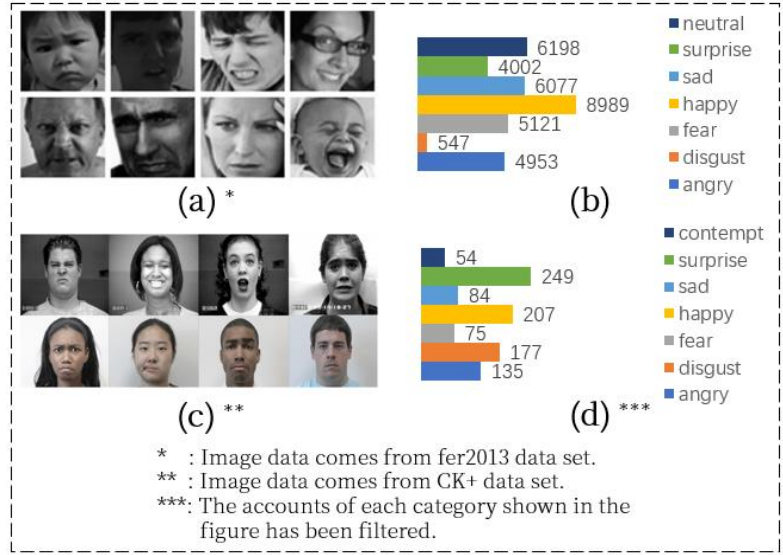

Fig. 2. (a) Part of Fer2013 data; (b) The number of Fer2013 categories; (c) Part of CK+ data;(d) The number of CK+ categories

## 2 Problem Formalization

In this part, we will introduce the scientific workflows of the neural network driven by three multimodal data used to recognize facial expressions.

### 2.1 Workflow of the Neural Network

The neural network requires a total of three multimodal data as input, which are images, facial landmarks and HOG features, and denote them as $I$, $L$ and $H$. After processing by the neural network, the three multimodal data will be mapped to $E$, representing the output of the neural network, where $E = (E_1, E_2, ..., E_i)$ and $E_i$ indicates the $i^{th}$ output of the network.

### 2.2 Details of the Workflow

When the neural network initially receives three multimodal data, it will first use the sub network to extract features:

$$F_1 = f_1(I) \tag{1}$$

$$F_2 = f_2(L) \tag{2}$$

$$F_3 = f_3(H) \qquad (3)$$

Where $F_1, F_2, F_3$ respectively denote the feature map extracted by three sub networks, $f_1, f_2, f_3$ denotes the processing of the input by the three sub networks. Then the neural network will fuse the three features into one, as shown in 3.1. The fused features contain data in multiple dimensions and will be sent to the classifier of the neural network for classification, as shown in (7) in 3.1. The working process of the neural network work can be summarized as:

$$E = classify\left(fuse(f_1(I), f_2(L), f_3(H))\right) \qquad (4)$$

In particular, each $E_i$ in $E$ corresponds to an expression label, and the final result, namely the expression of the face in the image can be determined by the following:

$$Expression = \max\{E_1, E_2, \ldots, E_i\} \qquad (5)$$

## 3 Network Structure

### 3.1 Overall Structure

The structure of neural network is shown in Fig.3, which includes three sub networks: "CNN" for processing face image, "LNN" for processing facial landmarks, and "HNN" for processing histogram of oriented gradient of image, as well as classifier for classification. Three sub networks work independently, and their structures are shown in Fig. 4.

The "preprocess" in Fig. 3 indicates the extraction of facial landmarks and HOG features from the input image. After feature extraction, the three sub networks are fused, and the three input features are connected along the axis to form a vector, as shown in "concatenate" in Fig. 3. Let $F_1, F_2, F_3$ respectively denote the feature map extracted by "CNN", "LNN" and "HNN", and $F_1 = [F_{1,1}, F_{1,2}, \ldots, F_{1,128}]$, $F_2 = [F_{2,1}, F_{2,2}, \ldots, F_{2,128}]$, $F_3 = [F_{3,1}, F_{3,2}, \ldots, F_{3,128}]$. First, global average pooling will be performed. For the fused feature $F$, $F \in R^{1 \times 384}$,

$$F = \frac{1}{128 \times 3} \sum_{i=1}^{3} \sum_{j=1}^{128} F_{i,j} \qquad (6)$$

The end of the neural network consists of two fully connected layers with 384 and 7 neurons, namely "FC" and "Expression" in Fig. 3 to process the fused feature $F$, which is first processed by the fully connected layer "FC" and the final result is

$$E_i = \frac{e^{z_i}}{\sum_{c=1}^{7} e^{z_c}} \qquad (7)$$

Where $E_i$ indicates the $i^{th}$ output of the network and $i = 1, 2, ..., 7$, $E_i \in [0, 1]$. Where $z_i$ indicates the output of the $i^{th}$ node in the fully connected layer "Expression".

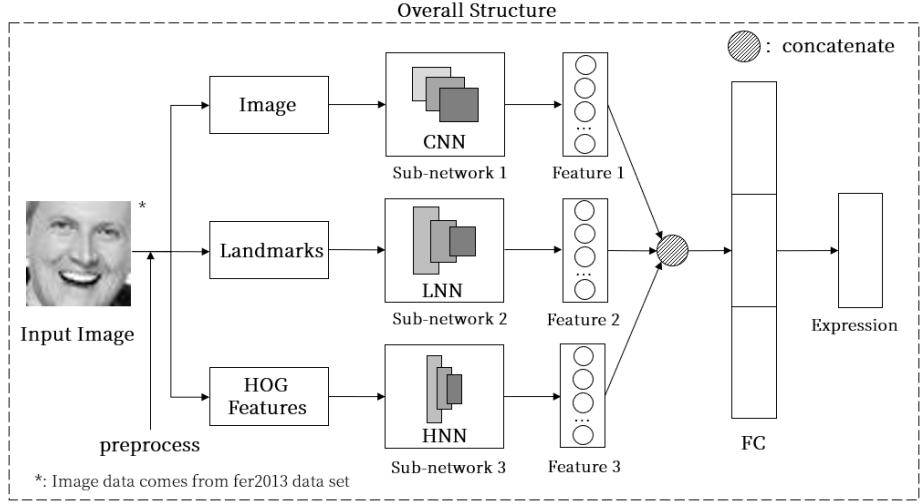

Fig. 3. Neural Network Architecture.

## 3.2 Sub network Structure

According to the CNN structure of the input image, as shown in "Neural Network for Image" in Fig. 4, the neural network extracts the deep semantic features of the facial image expression, which includes three convolution layers, three pooling layers, three batch normalization layers and three fully connected layers, and the input is the gray of 48x48 facial gray image. In order to steadily increase the depth of the network, the convolution filter in all layers is set to 3x3, which can effectively reduce the parameters and facilitate better extraction of deep features. The facial image processed by the neural network will be mapped to the feature map with the size of 128 through three fully connected layers with 4096, 1024 and 128 neurons (as shown by 'FCs' in Fig. 4), and then fused with the processing results of other multimodal data.

In order to improve the expression ability of facial image information by convolutional neural network, convolutional block attention module (CBAM) is added [12]. This module can effectively help the information flow in the network by learning which information needs to be emphasized or suppressed. After a feature map, the module will generate attention map along the two separate dimensions of channel and spatial. Let $F$ denotes the feature map and $F \in R^{C \times H \times W}$, the CBAM module will inference a one-dimensional channel attention map $M_c$ and a two-dimensional spatial attention map $M_s$,

$$F' = M_c(F) \otimes F \quad (8)$$

$$F'' = M_s(F') \otimes F' \quad (9)$$

where $M_c \in R^{C \times 1 \times 1}$ and $M_s \in R^{1 \times H \times W}$, $F''$ is the final output of the CBAM module.

Three CBAM modules are added to the convolutional neural network for facial image to assist feature extraction.

The LNN structure for processing facial landmarks is shown in Fig. 4 "Neural Network for Landmarks". The input is 68 facial landmarks, and because the facial landmarks can express a lot of information, in order to keep the original information as much as possible and make it cooperate with other multimodal data during fusion, the neural network is only composed of two fully connected layers with 1024, 128 neurons and maps the 68 landmarks of input to the feature map with the size of 128.

The HNN structure of histogram of oriented gradient is shown in "Neural Network for HOG" in Fig. 4, where the input is the extracted histogram of oriented gradient feature. Similar to the input face image, HOG features have other regions except the face, which will interfere with the work of the expression recognition model. Therefore, the neural network consists of three fully connected layers with 4096, 1024 and 128 neurons to process the input, so that it can focus on the face region without the interference of other regions in the image. The same as the above two multimodal data, the processed HOG features are mapped to the feature map with size of 128.

Batch normalization is carried out before the activation function of each layer of the neural network [13], which makes the input data distribution of each layer relatively stable to accelerate the model convergence. The use of the Dropout layer is particularly reduced in neural network to obtain comprehensive information of various dimensions.

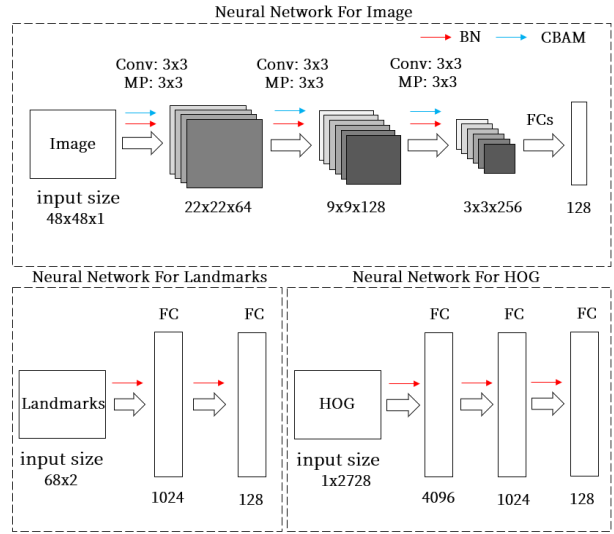

Fig. 4. Three Sub Network Structure; "FC" indicates fully connected layer; "FCs" indicates three fully connected layers with 4096, 1024 and 128 neurons.

For the three group of facial multimodal data, the neural network first uses three sub networks to extract features and map them to the feature map of size 128, and then carries out the fusion operation to establish a complete feature map. Now the feature map contains three kinds of different feature information, and finally outputs the prediction results through the full connection layer.

## 4 Training of Neural Network

### 4.1 Training Details

The training is based on the workstation equipped with AMD Ryzen 5600x processor and NVIDIA RTX 3080 graphics processor.

ReLU and softmax are used as the activation function in the sub network and activation function in the end classifier respectively. The neural network uses the cross entropy between the predicted value and the real value of facial expression as the loss function. N is the total number of samples, and θ represents the parameters of the network. $\hat{y}_i$ is the network prediction value of the $i^{th}$ sample, $y_i$ represents the real value, and $\hat{y}_i, y_i \in \{1, 2, ..., N\}$. The loss function of the network is as follows:

$$L(\theta) = -\frac{1}{N} \sum_{i=1}^{N} [y_i \log \hat{y}_i + (1 - y_i) \log(1 - \hat{y}_i)] \tag{10}$$

$$\hat{y}_i = f(X_i, \theta) \tag{11}$$

In the training process, Adam optimization algorithm [14] is used, and its parameters are set as alpha = 0.001, beta1 = 0.9, beta2 = 0.999 and epsilon= 10E-8, and the number of iteration is set as 500.

Because there are many data that are not aligned or non-face images in the Fer2013 dataset, and this will make the neural network unable to work normally. Therefore, this kind of noise data is first removed and then trained. The extraction of facial landmarks and hog features is based on Dlib [15]. The training data in Fer2013 and CK+ is expanded to 30 times of the original training data by random translation, rotation transformation and other data enhancement operations.

### 4.2 Experimental Results

The model is used to predict the test set for 10 times, and the average of the 10 prediction results is taken as the final result. The ROC curve of prediction results is shown in Fig. 5. In the Fer2013 data set, the overall accuracy of the model is 83.37%, the average accuracy of each category after individual test is 79.75%, and the accuracy of each category after individual test is angry-80.10%, dispute-53.04%, fear-70.79%, happy-98.48%, sad-78.26%, surprise-85.99%, neutral-91.60%, respectively. The confusion matrix is shown in Fig. 6. In the CK + dataset, the overall accuracy of the model is 99.41%, and the average accuracy of each category is 95.04%. The accuracy of each category is angry-100%, contempt-94.74%, dispute-95.16%, fear-96.23%, happy-100%, sadness-80.58%, and surprise-98.56%. The confusion matrix is shown in Fig. 6.

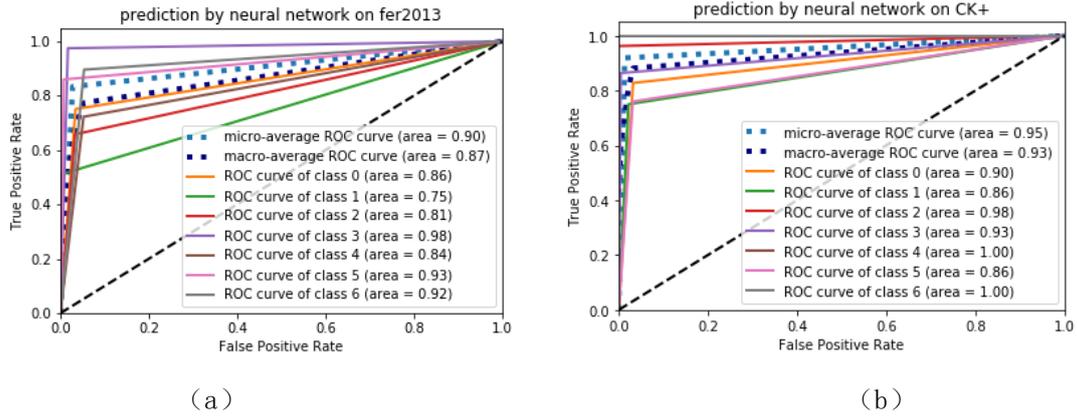

(a) (b)

Fig. 5. The performance of neural network on Fer2013 data set (a) and CK+ data set (b). (a) Class 0-6 corresponds to Angry, Disgust, Fear, Happy, Sad, Surprise and Neutral respectively; (b) Class 0-6 corresponds to Angry, Contempt, Disgust, Fear, Happy, Sadness and Surprise respectively.

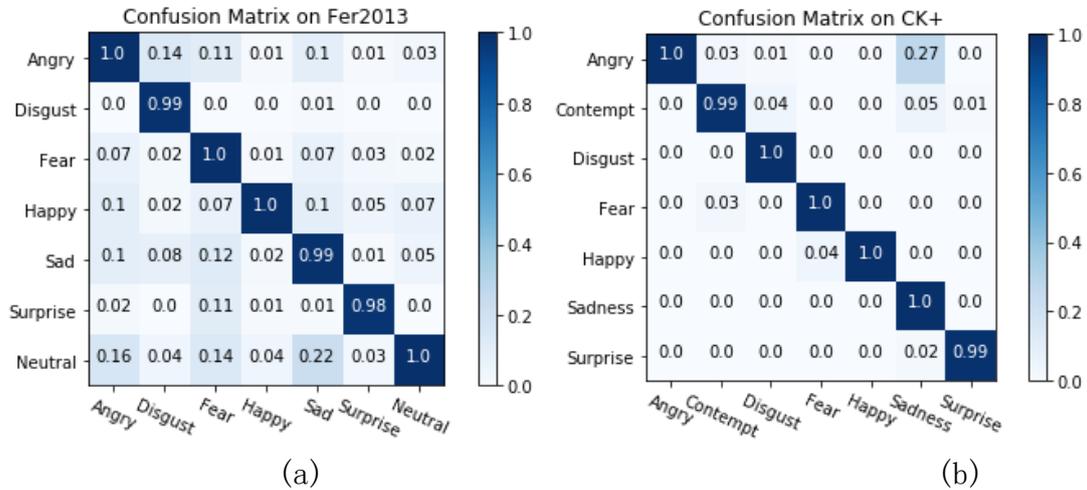

(a) (b)

Fig. 6. (a) The Confusion Matrix on the Fer2013 Dataset; (b) The Confusion Matrix on the CK + Dataset.

The performance of neural network on the processed (face alignment and noise elimination) Fer2013 test set is compared with other methods as shown in Table 1, in which [16] achieves an accuracy of 75.2% and 76.1% respectively on the original Fer2013 data set, and the face alignment data set in this experiment. [5] The two proposed convolution neural network models achieve an accuracy of over are 65% in the original data set. In this experiment, one of the models is tested on the processed data set, and the accuracy is 72.27%. And the experiment uses VGG [17] and ResNet [18] two kinds of neural network for comparison, the accuracy of the two is 74.36% and 73.53% respectively. The experimental results show that, in comparison with the other four neural networks, the facial multimodal data-driven neural network achieves better performance than other models due to the complementary of multimodal data.

On the processed CK+ data set, the neural network also achieved excellent performance. The comparison with other methods is shown in Table 2. The average accuracy of the two neural networks of VGG and ResNet are 92.48% and 91.27%, respectively, and the accuracy of classification using the SVM method is 87.17%;

among them [19] uses a deep convolutional neural network to construct a facial expression recognition system, including the Input Module, the Pre-processing Module, the Recognition Module and the Output Module, have achieved an accuracy of 80.303% on CK+. In the data set processed in this paper, the average accuracy has reached 89.51%.

Table 1. Comparison with other Methods on Fer2013 Data

| Labels | OURS | VGG | ResNet | [16] | [5] |
|---|---|---|---|---|---|
| Angry | 80.1% | 72.4% | 70.65% | 73.35% | 67.21% |
| Disgust | 53.04% | 49.6% | 52.13% | 51.06% | 48.34% |
| Fear | 70.79% | 66.81% | 64.21% | 65.52% | 63.5% |
| Happy | 98.48% | 94.32% | 94.24% | 94.28% | 90.39% |
| Sad | 78.26% | 69.1% | 68.41% | 75.92% | 71% |
| Surprise | 85.99% | 80.12% | 78.24% | 84.32% | 79.24% |
| Normal | 91.6% | 88.2% | 86.81% | 88.26% | 86.21% |
| Average | 79.75% | 74.36% | 73.53% | 76.10% | 72.27% |

Table 2. Comparison with other Methods on CK+ Data

| Labels | OURS | VGG | ResNet | SVM | [19] |
|---|---|---|---|---|---|
| Angry | 100% | 97.26% | 95.38% | 94.34% | 95.32% |
| Contempt | 94.74% | 90.35% | 90.06% | 88.28% | 89.2% |
| Disgust | 95.16% | 92.11% | 90.42% | 87.57% | 88.24% |
| Fear | 96.23% | 94.32% | 93.62% | 80.4% | 92.46% |
| Happy | 100% | 98.26% | 98.05% | 94.73% | 96.38% |
| Sadness | 80.58% | 78.8% | 75.43% | 74.31% | 72.23% |
| Surprise | 98.56% | 96.24% | 95.94% | 90.53% | 92.72% |
| Average | 95.04% | 92.48% | 91.27% | 87.17% | 89.51% |

In order to test the robustness of the neural network model, some noise data are added in the experiment, as shown in Fig. 7.

| | Original | Occlusion on mouth | Occlusion on eye | Original | Occlusion on mouth | High brightness |
|---|---|---|---|---|---|---|
| | 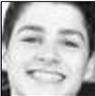 | 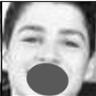 | 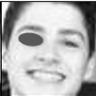 | 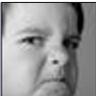 | 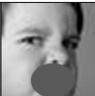 | 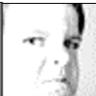 |
| OURS | 99.46% | 92.61% | 96.34% | 98.34% | 88.10% | 94.25% |
| VGG | 98.69% | 86.30% | 90.24% | 97.62% | 76.61% | 90.84% |
| ResNet | 99.29% | 82.15% | 87.52% | 98.08% | 72.38% | 91.27% |
| | Original | Occlusion on mouth | Occlusion on m&e | Original | Color distortion | Low brightness |
| | 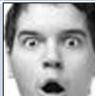 | 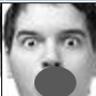 | 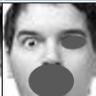 | 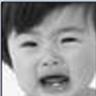 | 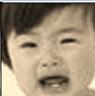 | 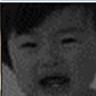 |
| OURS | 99.20% | 83.46% | 72.19% | 93.54% | 86.60% | 89.47% |
| VGG | 97.91% | 67.60% | NULL | 91.02% | 84.0% | 85.74% |
| ResNet | 98.39% | 63.27% | NULL | 90.88% | 83.31% | 85.20% |

\* : All the above image data comes from the fer2013 data set.
\*\* : The probability shown in the figure is the result that the neural network predicts, and 'NULL' indicates the prediction error of the neural network. 'm&e' indicates mouth and eye.

Fig. 7. Test of Neural Network under Interference.

The experimental results show that the neural network driven by three groups of facial multimodal data is better than the other two models not only in accuracy, but also in robustness. Compared with the other two models, when the face occlusion is small, the neural network can output high confidence results; when the face occlusion is large, the other two models barely predict correctly, and due to the participation of a number of multimodal data, the neural network based on multimodal data fusion can still predict the correct results. In the transformation of color and brightness, neural network still shows better accuracy than other models. It is worth noting that in many cases, VGG performs better than ResNet, and the latter has a deeper network layer. While the neural network proposed in this paper has fewer layers than both VGG and ResNet. This may indicate that the neural network with too much depth does not play a good role in improving the task of facial expression recognition, and the features extracted in this way are not suitable for the task.

**4.3 Experimental Conclusion**

The neural network expression recognition algorithm based on multimodal data fusion proposed in this paper has achieved good performance in a number of experimental tests, especially in the robustness. Compared with the current recognition model, the recognition accuracy has also been greatly improved, which has certain advantages.

**5. Conclusion**

In the field of facial expression recognition, this paper proposes a neural network recognition algorithm based on multimodal data fusion. By integrating the attention mechanism into the neural network, it makes full use of the complementarity among multimodal facial data to make the expression of facial image information more complete. The experimental results show that the algorithm has a great improvement in recognition accuracy and robustness. The disadvantage of the algorithm is that it needs more auxiliary information. In the follow-up research, we will try to combine the auxiliary information acquisition with the existing model to explore the fusion processing of more multimodal data information, so as to further achieve high-precision facial expression recognition in more complex scenarios.

DATA AVAILABILITY STATEMENT
The data that support the findings of this study are available from the corresponding author upon reasonable request.

Conflict of Interest
The authors declare no potential conflicts of interest with respect to the research, authorship, and/or publication of this article.

Acknowledgment

The authors received no financial support for the research, authorship, and/or publication of this article.Reference
[1] Sariyanidi E, Gunes H, Cavallaro A. Automatic analysis of facial affect: A survey of registration, representation, and recognition[J]. IEEE transactions on pattern analysis and machine intelligence, 2014, 37(6): 1113-1133.
[2] Lekshmi P V, Sasikumar M. Analysis of facial expression using Gabor and SVM[J]. International Journal of Recent Trends in Engineering, 2009, 1(2): 47.
[3] Lanitis A, Taylor C J, Cootes T F. Automatic interpretation and coding of face images using flexible models[J]. IEEE Transactions on Pattern Analysis and machine intelligence, 1997, 19(7): 743-756.
[4] Tang Y. Deep learning using support vector machines[J]. CoRR, abs/1306.0239, 2013, 2.
[5] Agrawal A, Mittal N. Using CNN for facial expression recognition: a study of the effects of kernel size and number of filters on accuracy[J]. The Visual Computer, 2020, 36(2): 405-412.
[6] LI Yong, LIN Xiao-Zhu, JIANG Meng-Ying. Facial Expression Recognition with Cross-connect LeNet-5 Network. Acta Automatica Sinica, 2018, 44(1): 176-182. doi: 10.16383/j.aas.2018.c160835
[7] Tanaka J W, Gordon I. Features, configuration, and holistic face processing[J]. The Oxford handbook of face perception, 2011: 177-194.
[8] Chen H, Hu N, Cheng Z, et al. A deep convolutional neural network based fusion method of two-direction vibration signal data for health state identification of planetary gearboxes[J]. Measurement, 2019, 146: 268-278.
[9] Du S, Li T, Gong X, et al. A hybrid method for traffic flow forecasting using multimodal deep learning[J]. arXiv preprint arXiv:1803.02099, 2018.
[10] Goodfellow I J, Erhan D, Carrier P L, et al. Challenges in representation learning: A report on three machine learning contests[C]//International conference on neural information processing. Springer, Berlin, Heidelberg, 2013: 117-124.
[11] Lucey P, Cohn J F, Kanade T, et al. The extended cohn-kanade dataset (ck+): A complete dataset for action unit and emotion-specified expression[C]//2010 ieee computer society conference on computer vision and pattern recognition-workshops. IEEE, 2010: 94-101.
[12] Woo S, Park J, Lee J Y, et al. Cbam: Convolutional block attention module[C]//Proceedings of the European conference on computer vision (ECCV). 2018: 3-19.
[13] Ioffe S, Szegedy C. Batch normalization: Accelerating deep network training by reducing internal covariate shift[C]//International conference on machine learning. PMLR, 2015: 448-456.
[14] Kingma D P, Ba J. Adam: A method for stochastic optimization[J]. arXiv preprint arXiv:1412.6980, 2014.
[15] King D E. Dlib-ml: A machine learning toolkit[J]. The Journal of Machine Learning Research, 2009, 10: 1755-1758.